  \documentclass[times,twocolumn, final ,authoryear]{elsarticle}
  
  \usepackage{ycviu_arxiv}
  \usepackage{framed,multirow}

    \usepackage{amsmath,amssymb}
    \usepackage{booktabs}
    \usepackage{tablefootnote}
    \usepackage{latexsym}
    \usepackage{array}
    \usepackage{xspace}
    \usepackage{threeparttable}
    
    \usepackage{url}
    \usepackage{xcolor}
    \usepackage[nolist,nohyperlinks]{acronym}
    
    \usepackage[capitalise, noabbrev]{cleveref}
    \usepackage{pdfpages} 
    \definecolor{newcolor}{rgb}{.8,.349,.1}
    
    \graphicspath{{/}}
    \DeclareMathOperator*{\argmax}{arg\,max}
    
    \newcommand{\R}{\mathbb{R}}
    \newcommand*{\eg}{e.g.\@\xspace}

  \makeatletter
  \DeclareRobustCommand*\textsubscript[1]{%
  	\@textsubscript{\selectfont#1}}
  \def\@textsubscript#1{%
  	{\m@th\ensuremath{_{\mbox{\fontsize\sf@size\z@#1}}}}}
  \makeatother
  
  \usepackage{pifont}
  \newcommand{\xmark}{\ding{55}}%
  
  \journal{Computer Vision and Image Understanding}
  
\begin{document}

  	\ifpreprint
  	\setcounter{page}{1}
  	\else
  	\setcounter{page}{1}
  	\fi
  	
  	\begin{frontmatter}
  		
  		\title{DRAU: Dual Recurrent Attention Units for Visual Question Answering}
  		
  		\author[add1]{Ahmed \snm{Osman}\corref{cor1}}
  		\ead{ahmed.osman@hhi.fraunhofer.de}
  		\author[add1]{Wojciech \snm{Samek}\corref{cor1}}
  		\ead{wojciech.samek@hhi.fraunhofer.de}

		\address[add1]{Fraunhofer Heinrich Hertz Institute, Einsteinufer 37, Berlin 10587, Germany}
  		
	\begin{acronym}[JSONP]\itemsep0pt 
		\acro{VQA}{Visual Question Answering}
		\acro{RAU}{Recurrent Attention Unit}
		\acro{DRAU}{Dual Recurrent Attention Units}
		\acro{RTAU}{Recurrent Textual Attention Unit}
		\acro{RVAU}{Recurrent Visual Attention Unit}
		\acro{CNN}{convolutional neural network}
		\acro{RNN}{recurrent neural network}
		\acro{GRU}{gated recurrent unit}
		\acro{LSTM}{long short-term memory}
		\acro{DCA}{Dual Convolution Attention}
	\end{acronym}

	\begin{abstract}
		\ac{VQA} requires AI models to comprehend data in two domains, vision and text. Current state-of-the-art models use learned attention mechanisms to extract relevant information from the input domains to answer a certain question. Thus, robust attention mechanisms are essential for powerful VQA models. In this paper, we propose a recurrent attention mechanism and show its benefits compared to the traditional convolutional approach.  We perform two ablation studies to evaluate recurrent attention. First, we introduce a baseline \ac{VQA} model with visual attention and test the performance difference between convolutional and recurrent attention on the VQA 2.0 dataset. Secondly, we design an architecture for VQA which utilizes dual (textual and visual) \acp{RAU}. Using this model, we show the effect of all possible combinations of recurrent and convolutional dual attention. Our single model outperforms the first place winner on the VQA 2016 challenge and to the best of our knowledge, it is the second best performing single model on the VQA 1.0 dataset. Furthermore, our model noticeably improves upon the winner of the VQA 2017 challenge. Moreover, we experiment replacing attention mechanisms in state-of-the-art models with our \acp{RAU} and show increased performance. 
		
	\end{abstract}
  		
  \end{frontmatter}

  \section{Introduction}
  \label{sec1}
  Although \acp{CNN} and \acp{RNN}
  have been successfully applied to various image and natural language processing tasks (cf.\ \cite{he2015, BosTIP18, bahdanau2014, nallapati2016abstractive}), these breakthroughs only slowly translate to multimodal tasks such as \ac{VQA} where the model needs to create a joint understanding of the image and question. Such multimodal tasks require designing highly expressive joint visual and textual representations. On the other hand, a highly discriminative multi-modal feature fusion method is not sufficient for all VQA questions, since global features can contain noisy information for answering questions pertaining to certain local parts of the input. This motivation has led to the use of attention mechanisms in VQA.

  Attention mechanisms have been extensively used in VQA recently \citep{anderson2017, fukui2016, kim2016b}. They attempt to make the model selectively predict based on segments of the spatial or lingual context. However, most attention mechanisms used in VQA models are rather simple, consisting of two convolutional layers and a softmax function to generate the attention weights which are summed over the input features. These shallow attention mechanisms could fail to select the relevant information from the joint representation of the question and image for complex questions. According to the literature in human cognition, humans process cognitive attention both spatially and temporally \citep{rensink2000}. Consequently, recurrent attention mechanisms come to mind. Creating attention for complex questions, particularly sequential or relational reasoning questions, requires processing information in a sequential manner which recurrent layers are better suited due to their intrinsic ability to capture relevant information over an input sequence.

  In this paper, we propose a \ac{RNN}-based attention mechanism for visual and textual attention. We argue that embedding an RNN in the attention mechanism helps the model process information in a sequential manner and determine what is relevant to solve the task. We refer to the combination of RNN embedding and attention as  \ac{RTAU} and \ac{RVAU} respective of their purpose. Furthermore, we employ these units in a fairly simple network, referred to as \ac{DRAU} network, and show competitive results compared to state-of-the-art models. 
  
  Our main contributions are the following:
  \begin{itemize}
  	\item We introduce a novel approach to generate soft attention for VQA. To the best of our knowledge, this is the first attempt to generate attention maps using recurrent neural networks in the VQA domain.
  	
  	\item We conduct a direct comparison between two identical models except for their attention mechanism. In this controlled environment, the recurrent attention outperforms the convolutional attention significantly (4\% absolute difference). 
  	
  	\item We propose a network that utilizes two \acp{RAU} to co-attend the multi-modal inputs. We perform an ablation study to further test the effect of recurrent attention in our model. The results show a significant improvement when using both \acp{RAU} compared to using convolutional attention.
  	
  	\item   \acp{RAU} are modular, thus, they can substitute existing attention mechanisms in most models fairly easily. We show that state-of-the-art models with RVAU or RTAU ``plugged-in" perform consistently better than their standard counterparts.
  	
  	\item  We show that our network outperforms the  VQA 2016 and 2017 challenge winners and performs close to the current state-of-the-art single models. Additionally, we provide qualitative results showing subjective improvements over the default attention used in most VQA models.
  \end{itemize}
  
  In Section 2, we review related work for recurrent attention and VQA methods. In Section 3, we break down the components of the DRAU network and explain the details of a \ac{RAU}. In Section 4, we compare convolutional and recurrent attention in a baseline model, conduct ablation experiments using \ac{DRAU}, and report the results of substituting attention mechanisms of state-of-the-art models with \acp{RAU} on the VQA 2.0 dataset \citep{goyal2016}. Then we compare our model against the state-of-the-art on the VQA 1.0 \citep{antol2015} and 2.0 datasets. In Section 5, we compare the difference in attention maps between standard and recurrent attention with qualitative examples to illustrate the effect of RAUs. Finally, we conclude the paper in Section 6 and discuss future work.

  \section{Related Work}
  This section discusses related recurrent attention mechanisms and common methods that have been explored in the past for VQA.
  
  	\paragraph{Recurrent Attention}
  	\ac{RNN}-based attention models have been used outside of the VQA domain.
  	\cite{mnih2014} train a recurrent neural network model for digit recognition. Their model selects a
  	sequence of regions of an image and processes the selected regions at high
  	resolutions to save computational power. However, the whole architecture is a monolithic
  	recurrent model that limits its use in existing VQA models since it can not easily substitute VQA attention mechanisms. Moreover, the model
  	is not differentiable and requires training with reinforcement learning which is highly inefficient to train in
  	practice due to training instability \citep{lanctot2017} and complexity of reward function
  	design \citep{paulus2017}.
  	Closer to our work, \cite{homayounfar2018} uses a recurrent attention mechanism to attend to lane boundaries for traffic lane counting. There exists some differences compared to our attention unit. Their attention mechanism uses multi-step training to train
  	the attention unit while \acp{RAU} do not require any attention-specific training. It is worth noting that \cite{homayounfar2018} use a vanilla Convolutional RNN for attention. Furthermore, they sample the input feature maps from different scales to provide information from different granularites. Trying different recurrent architectures like GRU \citep{cho2014a},
  	Grid-LSTM \citep{kalchbrenner2015}, and Conv-LSTM \citep{shi2015} and their effect on attention could be an interesting research direction in the future.

  \paragraph{Bilinear pooling representations}\citet{fukui2016} use compact bilinear pooling to attend over the image features and combine it with the language representation. The basic concept behind compact bilinear pooling is approximating
  the outer product by randomly projecting the embeddings to a higher
  dimensional space using Count Sketch projection \citep{charikar2004} and then exploiting
  Fast Fourier Transforms to compute an efficient convolution. An ensemble model using MCB won first place in VQA (1.0) 2016 challenge.
  \citet{kim2016b} argues that compact bilinear pooling is still expensive
  to compute and shows that it can be replaced by element-wise product (Hadamard product) and a linear mapping (i.e.\ fully-connected layer) which gives a lower dimensional representation and also improves the model accuracy.
  \citet{ben-younes2017} proposed using Tucker decomposition \citep{tucker1966} with a low-rank matrix constraint as a bilinear representation. \cite{yu2017a} utilize matrix factorization tricks to create a multi-modal factorized bilinear pooling method (MFB). Later, \cite{yu2017} generalizes the factorization for higher-order factorized pooling (MFH).
  
  \paragraph{Attention-based}
  \cite{xu2015} propose a memory network that predicts visual attention based on a dot product between image features and word embedding. They utilize a separate embedding that predicts visual ``evidence'' related to the question. The evidence embedding can be computed iteratively (2-hops) to collect more evidence.
  \citet{lu2016} were the first to feature
  a co-attention mechanism that applies attention to both the question
  and image. \citet{nam2016b} use a Dual Attention Network (DAN) that
  employs attention on both text and visual features iteratively to predict the result. The goal behind this is to allow the image and question attentions to guide each other in a co-dependent manner. \cite{schwartz2017} use high-order correlations between the multimodal input and multiple-choice answers to guide the model's attention. It is worth noting that this model processes attention not only for both the question and image, but also for the answer. However, their VQA model utility is gravely weakened by their dependence on multiple-choice answer embedding which renders their model computationally infeasible for standard open-ended VQA tasks.
  \paragraph{\acp{RNN} for VQA}
  Using \acp{RNN} for VQA models has been explored in the past, but, to the best of our knowledge, has never been used as an attention mechanism. \citet{xiong2016a} build upon the dynamic memory network from \citet{kumar2015} and proposes DMN+. DMN+ uses episodic modules which contain attention-based GRUs. Note that this is not the same as what we propose; \citeauthor{xiong2016a} generate soft attention using \textit{convolutional layers} and uses the output to substitute the update gate of the GRU. In contrast, our approach uses the \textit{recurrent layers} to explicitly generate the attention. \citet{noh2016c} propose recurrent answering units in which each unit is a complete module that can answer a question about an image. They use joint loss minimization to train the units. However during testing, they use the first answering unit which was trained from other units through backpropagation.

  \section{Dual Recurrent Attention in VQA}
\begin{figure*}
  	\centering
  	\includegraphics[trim=0cm 1cm 0cm 0cm, width=0.9\linewidth]{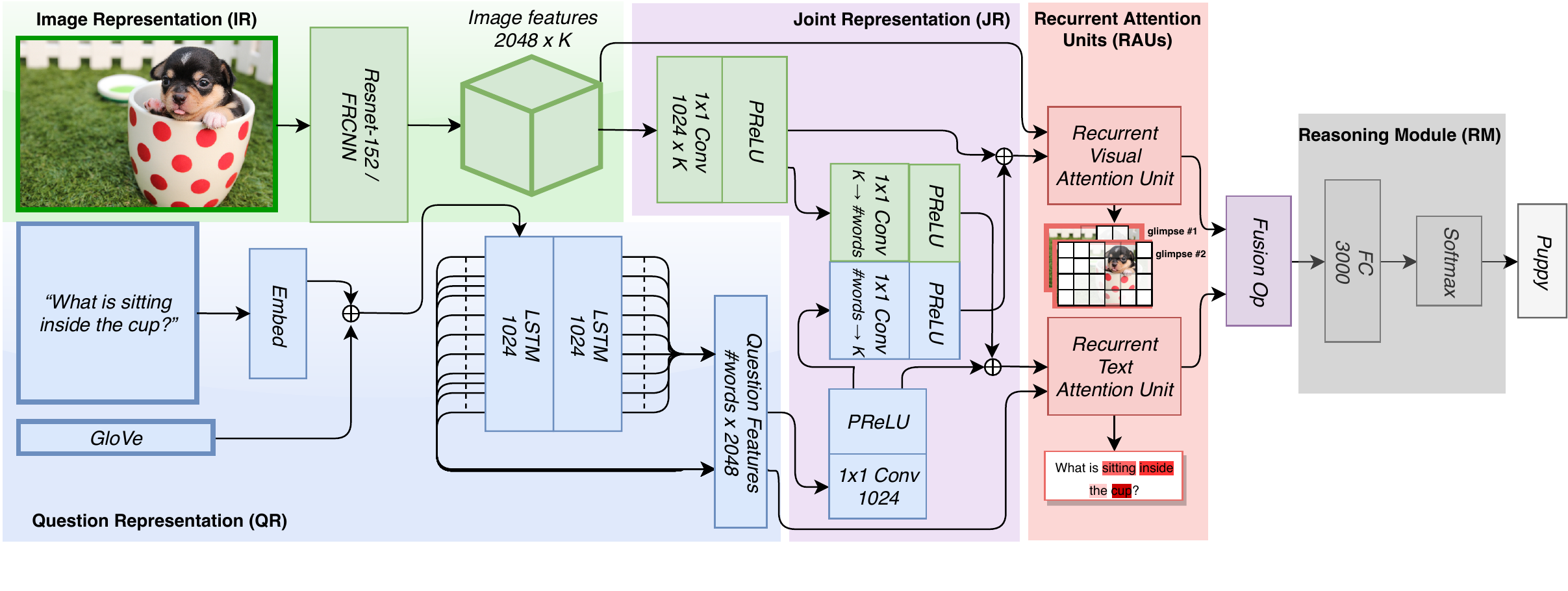}
  	\caption{The proposed network. $\oplus$ denotes concatenation.}
  	\label{fig:vqanet}
  \end{figure*}

  In this section, we define our attention mechanism. Then, we describe the components of our VQA model in this section. All modules are annotated in \cref{fig:vqanet} for reference.
  
  \begin{figure}
  	\centering
  	\includegraphics[width=0.9\linewidth]{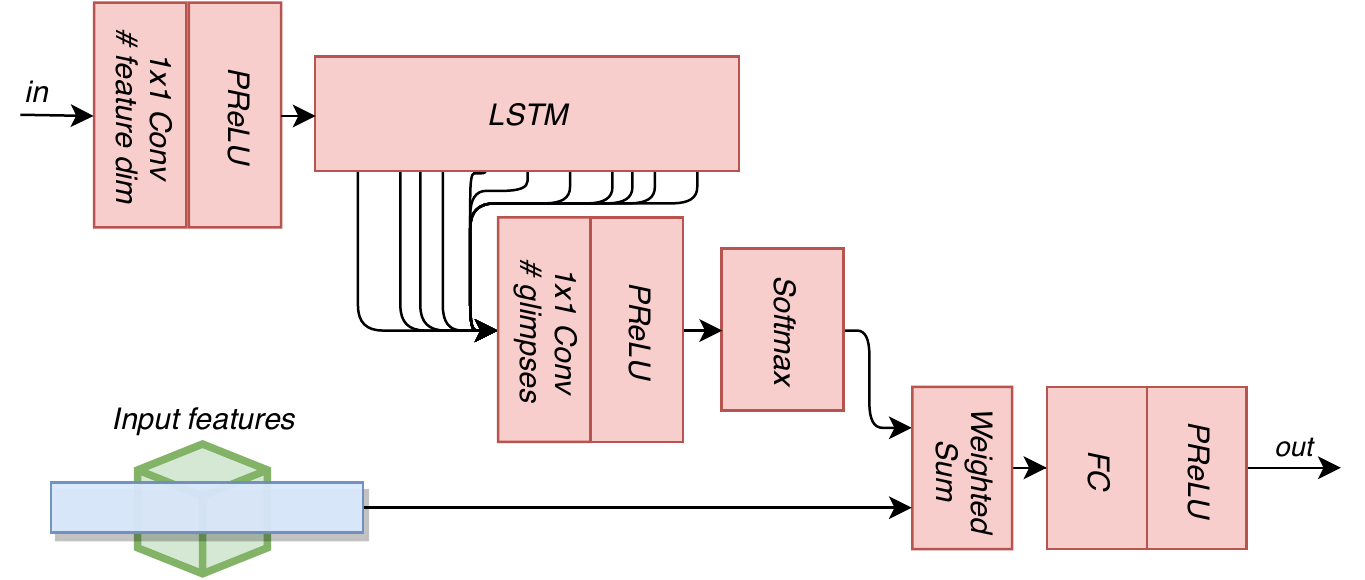}
  	\caption{Recurrent Attention Unit.}
  	\label{fig:rau}
  \end{figure}
  
  \subsection{\acfp{RAU}}
  The \ac{RAU} receives a multi-modal multi-channel representation of the inputs, $ X$. To scale down the input representation, a RAU starts with a $1\times1$ convolution and PReLU activation \citep{he2015}:
  \begin{equation}
  \begin{gathered}
  c_{a} = \mathrm{PReLU}\ \big(W_a\ X\big),\\
  X \in \R^{K \times \phi}
  \end{gathered}
  \end{equation}
  where $W_a$ is the $1\times1$ convolution weights, $X$ is the multimodal input to the RAU, $K$ is the shape of the target attention (\eg image pixels/number of visual objects or question length), and $\phi$ is the number of channels in the input.
  
  Furthermore, we feed the previous output into a unidirectional LSTM:
  \begin{equation}
  h_{a,n} = \mathrm{LSTM}\ \big(c_{a,n}\big)
  \end{equation}
  where $h_{a,n}$ is the hidden state at time $n$. Each hidden state processes the joint features at each location/word of the input and decides which information should be kept and propagated forward and which information should be ignored.
  
  To generate the attention weights, we feed all the hidden states of the previous LSTM to a $1\times1$ convolution layer followed by a softmax function. The $1\times1$ convolution layer could be interpreted as the {\it number of glimpses} the model sees.
  \begin{equation}
  W_{att,n} = \mathrm{softmax}\ \Big(\mathrm{PReLU}\ \big(W_g\ h_{a,n}\big)\Big)
  \end{equation}
  where $W_g$ is the glimpses' weights and $W_{att,n}$ is the attention weight vector.
  
  Next, we use the attention weights to compute a weighted average of the image and question features.
  \begin{equation}
  att_{a,n} = \sum_{n = 1}^{N} W_{att,n}\ f_n
  \end{equation}
  where $f_n$ is the input representation and $att_{a,n}$ is the attention applied on the input.
  Finally, the attention maps are fed into a fully-connected layer and a PReLU activation. \cref{fig:rau} illustrates the structure of a RAU.
  \begin{equation}
  y_{att,n} = \mathrm{PReLU}\ \big(W_{out}\ att_{a,n}\big)
  \end{equation}
  where $W_{out}$ is a weight vector of the fully connected layer and $y_{att,n}$ is the output of the RAU.

\subsection{Input Representation}\label{sub:input}
\paragraph{Image representation (IR)} In our baseline and ablation studies, we use two types of image representations. First, a 152-layer ``ResNet" pretrained CNN from \citet{he2015} to extract image features. Similar to \citep{fukui2016, nam2016b}, we resize the images to $448\times 448$ and extract the last layer before the final pooling layer (res5c) with size $2048\times14\times14$. Finally, we use $l_2$ normalization  on all dimensions.
	Recently, \citet{anderson2017} have shown that object-level features can provide a significant performance uplift compared to global-level features from pretrained CNNs. Therefore, we use Faster R-CNN features \citep{ren2015b} with a fixed number of proposals per image ($K = 36$) for the DRAU model and its variants.
\paragraph{Question representation (QR)}
We use a fairly similar representation as \cite{fukui2016}. In short, the question is tokenized and encoded using an embedding layer followed by a $tanh$ activation. We also exploit pretrained GloVe vectors \citep{pennington2014glove} and concatenate them with the output of the embedding layer. The concatenated vector is fed to a two-layer unidirectional LSTM that contains 1024 hidden states each. In contrast to \citeauthor{fukui2016}, we use all the hidden states of both LSTMs rather than concatenating the final states to represent the final question representation.

\subsection{Use of $1\times1$ Convolution and PReLU}
We apply multiple $1\times1$ convolution layers in the network for mainly two reasons. First, they learn weights from the image and question representations in the early layers. This is important especially for the image representation, since it was originally trained for a different task. This is also true for the question representation to a lesser degree (GloVe vectors are trained on co-occurrence statistics) Second, they are used to generate a common representation size. To obtain a joint representation (JR), we apply $1\times1$ convolutions followed by PReLU activations on both the image and question representations. Through empirical evidence, PReLU activations were found to reduce training time significantly and improve performance compared to ReLU and $tanh$ activations. 

\subsection{Fusion operation}
A fusion operation is used to merge the textual and visual branches. For DRAU, we use MCB \citep{fukui2016, gao2016}.

\subsection{Reasoning module (RM)}
The result of the fusion is given to a many-class classifier using the top 3000 frequent answers. We use a single-layer softmax with cross-entropy loss. This can be written as:
\begin{equation}
P_a = \mathrm{softmax}\Big(\mathrm{fusion\_op}\big(y_{text}, y_{vis}\big) W_{ans} \Big)
\end{equation}

where $y_{text}$ and $y_{vis}$ are the outputs of the RAUs, $W_{ans}$ represents the weights of the multi-way classifier, and $P_a$ is the probability of the top 3000 frequent answers.

The final answer $\hat{a}$ is chosen according to the following:
\begin{equation} \hat{a} = \argmax P_a \end{equation}

\section{Experiments and Results}\label{sec:experiments}
Experiments are performed on the VQA 1.0 and 2.0 datasets \citep{goyal2016,antol2015}. These datasets use images from the MS-COCO dataset \citep{lin2014} and generate questions and labels (10 labels per question) using Amazon's Mechanical Turk. Compared to VQA 1.0, VQA 2.0 adds more image-question pairs to balance the language prior present in the VQA 1.0 dataset \citep{goyal2016}. The ground truth answers in the VQA dataset are evaluated using human consensus:
\begin{equation}
\mathrm{Acc}(a) = \min \bigg(\frac{\sum{a \text{ is in human annotation} }}{3}, 1\bigg)
\end{equation}

We evaluate our results on the \textit{validation}, \textit{test-dev}, \textit{test-std} splits of each dataset. We test on multiple splits due to the fact that submissions to the test server are limited (for test and test-dev splits). Thus, we mainly evaluate models on the validation split unless comparing results with the state-of-the-art.

To train our model, we use Adam \citep{kingma2014} for optimization with $\beta_1= 0.9 \text{, } \beta_2 = 0.999$, and  an initial learning rate of
$\epsilon= 7\times10^{-4}$. The models are trained with a small batch size of 32 for  400K iterations. We did not fully explore tuning the batch size which explains the relatively high number of training iterations.
Dropout $(p = 0.3)$ is applied after each LSTM and after the fusion operation. All weights are initialized as described in \citep{glorot2010} except LSTM layers which use an uniform weight distribution.

Since VQA datasets provide 10 answers per image-question pair, we randomly sample one answer at each training iteration.

\begin{table*}
	\center
	\caption{Simple Net model and ablation study  of the DRAU model results on the VQA 2.0 \textit{validation} split. The number of trainable parameters are shown for each model. IF indicates the type of image features the model uses. VG indicates whether the method uses external data augmentation from the Visual Genome dataset.}
	
		\begin{tabular}{lcccccccc}
			\toprule
			\multicolumn{9}{c}{VQA 2.0 Validation Split} \\
			\midrule
			Model                           & \# Parameters     & IF     & Fusion & VG         & All                & Y/N   & Num.  & Other \\
			\cmidrule(r){1-1}       \cmidrule(l){2-5}  \cmidrule(l){6-9}
			Simple Net (Conv. Visual Attn.) & $37.4\times 10^9$ & ResNet & \xmark & \checkmark & 41.06              & 66.01 & 28.08 & 25.51 \\ %
			Simple Net w/RVAU               & $37.3\times10^9$  & ResNet & \xmark & \checkmark & 45.12              & 66.24 & 28.48 & 33.46 \\
			\midrule	
			DCA (Conv. Attn)                & $138.4\times10^9$ & FRCNN  & MCB    & \xmark     & $59.42 \pm 0.09$   & 77.80 & 36.28 & 51.57 \\ %
			DCA w/RVAU                      & $138.4\times10^9$ & FRCNN  & MCB    & \xmark     & $60.90 \pm 0.02$   & 79.04 & 39.63 & 52.73 \\ %
			DCA w/RTAU                      & $138.4\times10^9$ & FRCNN  & MCB    & \xmark     & $ 58.76 \pm 0.08 $ & 77.56 & 35.54 & 50.61 \\
			DRAU                            & $138.4\times10^9$ & FRCNN  & MCB    & \xmark     & $61.36 \pm 0.02$   & 79.74 & 40.03 & 53.03 \\
			\bottomrule
		\end{tabular}
	
	\label{tab:vqa_dau_baselines}
\end{table*}

\subsection{Attention ablation studies}

\paragraph{Convolutional versus recurrent attention}


\begin{figure}
	\centering
	\includegraphics[width=0.8\linewidth]{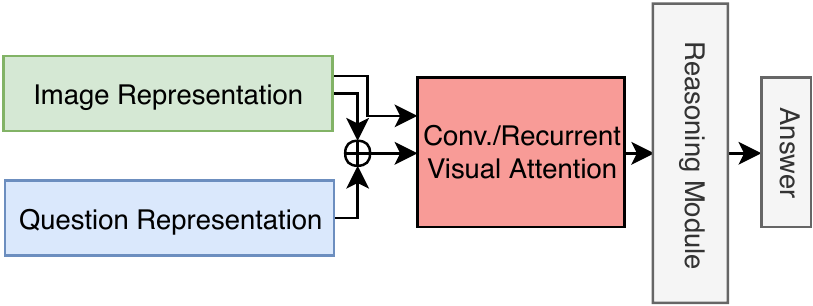}
	\caption{Simple Net architecture. This model uses components from the DRAU model (\cref{fig:vqanet}) to be used as a baseline for attention evaluation. Multimodal features are concatenated ($\oplus$) and fed directly to the attention mechanism.}
	\label{fig:simplenet]}
\end{figure}


We compare using convolution against our recurrent attention in a simple baseline model. 
This baseline VQA model uses the same question representation as described in the previous section and the ResNet global image features. The input features are simply concatenated and sent to a visual attention mechanism. The processed attention is fed to the reasoning module. We refer to this model as \textit{Simple Net} (illustrated in \cref{fig:simplenet]}). Simple Net was trained twice: once with convolutional attention and once with \ac{RVAU}. To avoid any type of parameter advantage, both models were designed to have the same number of parameters approximately. Both baselines use the \textit{train} split and Visual Genome \citep{krishna2017} for training, and were evaluated on the VQA 2.0 validation split. The results of Simple Net in \cref{tab:vqa_dau_baselines} show a clear advantage of recurrent attention outperforming convolutional attention by over 4\% absolute overall accuracy.

In second ablation study, we aim to examine the effect of different combinations of recurrent attention by training 4 different variants of the \ac{DRAU} model that only differ in the type of attention used. The models are the following: 
	\begin{itemize}
		\item \ac{DCA}: DRAU with dual convolutional attention, i.e.\ text att.(Convolution) -- image att.(Convolution).
		\item \ac{DCA} w/\ac{RVAU}: text att.(Convolution) -- image att.(\ac{RVAU}).
		\item \ac{DCA} w/\ac{RTAU}: text att.(\ac{RTAU}) -- image att.(Convolution).
		\item DRAU: text att.(\ac{RTAU}) -- image att.(\ac{RVAU}).
	\end{itemize}

	Similar to the baseline models, we match the number of parameters in the attention units such that all models have roughly the same number of parameters to ensure that differences are not from a higher number of parameters. All models were trained using Faster R-CNN image features on the \textit{train} split. Additionally, we train each variant with 3 different initial seeds to
	show the effect of parameter initialization.

The results in \cref{tab:vqa_dau_baselines}  show the evaluation results of this ablation study on the VQA 2.0 validation split. We report the mean and standard deviation of the each model. Using RVAU improves
	the baseline \ac{DCA} model by 1.48\%. While swapping convolutional text attention with
	RTAU reduces performance by 0.66\%, using both RTAU and RVAU in the
	same model improves the overall performance by almost 2\%. This might
	indicate that multimodal recurrent text attention thrives with certain architectural designs since it improves performance in our DRAU model.

In conclusion, using \ac{RVAU} significantly improves network accuracy. \ac{RTAU} requires more careful use, but in an appropriate model, improves performance significantly.

\subsection{Using \acp{RAU} in other models}

To verify the effectiveness of the recurrent attention units, we replace the convolutional attention layers in MCB \citep{fukui2016} and MUTAN \citep{ben-younes2017} with RVAU (visual attention). Additionally, we replace the textual attention in MFH \citep{yu2017} with recurrent attention.

For MCB we remove all the layers after the first MCB operation until the first 2048-d output and replace them with RVAU. Due to GPU memory constraints, we reduced the size of each hidden unit in RVAU's LSTM from 2048 to 1024. In the same setting, RVAU significantly helps improve the original MCB model's accuracy as shown in  \cref{tab:mcb_rvau}.

Furthermore, we test RVAU in the MUTAN model. The authors use a multimodal vector with dimension size of $510$ for the joint representations. For coherence, we change the usual dimension size in RVAU to $510$. At the time of this writing, the authors have not released results on VQA 2.0 using a single model rather than a model ensemble. Therefore, we train a single-model MUTAN using the authors' implementation.\footnote{\url{https://github.com/Cadene/vqa.pytorch}} The story does not change here, RVAU improves the model's overall accuracy.

Finally, we replace the convolution text attention in MFH with RTAU (text attention). We train two networks, the standard MFH network and MFH with RTAU, on the VQA 2.0 train split and test on the validation split. It is apparent that RTAU improves the overall accuracy of MFH from \cref{tab:mcb_rvau}. Note that the text attention in MFH is ``self-attending'' which means that the textual attention does not
	interact with visual content in this setting. This is different from our DRAU model where RTAU uses a joint representation of the question and image to predict the textual attention. This gives insight about the difference of performance between the ablation study and the modified MFH results. While the performance improvement might not look large for some models, it is consistent which shows that RAUs can reliably improve existing state-of-the-art models with different architectures.

\begin{table}[tbph]
	\center
	\begin{threeparttable}
		
		\caption{Results of state-of-the-art models with RAUs.}
		\begin{tabular}[t]{lcccc}
			\toprule
			
			\multicolumn{5}{c}{VQA 2.0 Test-dev Split} \\
			\midrule
			Model                           & All   & Y/N   & Num.  & Other \\
			\cmidrule(r){1-1}       \cmidrule(l){2-2}  \cmidrule(l){3-5}
			MCB \citep{fukui2016} \tnote{2} & 61.96 & 78.41 & 38.81 & 53.23 \\ %
			MCB w/RVAU                      & 62.33 & 77.31 & 40.12 & 54.64 \\
			\midrule
			MUTAN (Ben. et al.,2017)        & 62.36 & 79.06 & 38.95 & 53.46 \\ %
			MUTAN w/RVAU                    & 62.45 & 79.33 & 39.48 & 53.28 \\
			\bottomrule
			\multicolumn{5}{c}{VQA 2.0 Validation Split} \\
			
			\cmidrule(r){1-1}       \cmidrule(l){2-2}  \cmidrule(l){3-5}
			MFH  \citep{yu2017}             & 64.31 & 82.26 & 43.49 & 56.17 \\ %
			MFH w/RTAU                      & 64.38 & 82.35 & 43.31 & 56.3  \\

			\bottomrule
		\end{tabular}
		\begin{tablenotes}
			\item[2] \footnotesize \url{http://www.visualqa.org/roe_2017.html}
		\end{tablenotes}
	\end{threeparttable}
	\vspace{0.3cm}
	\label{tab:mcb_rvau}
\end{table}

  \subsection{DRAU versus the state-of-the-art}
  
  In this section, we present the results of our model in the scope of state-of-the-art models. Although our model does not utilize some extra knowledge and performance optimization techniques (Visual Genome augmentation, model ensembles, hyperparameter tuning), we choose to show models that do so in order to present our results from a practical point of view. 
  
  \paragraph{VQA 1.0} \cref{tab:vqa1_test} shows a comparison between DRAU and other state-of-the-art models. Excluding model ensembles, DRAU performs favorably against other models. To the best of our knowledge, \cite{yu2017} has the best reported single model performance of $67.5\%$ on the \textit{test-std} split. Our single model (DRAU) comes a very close second to the current state-of-the-art single model.

  \begin{table*}[t]
  	\center
  	\caption[DRAU compared to the state-of-the-art on the VQA 1.0 dataset.]{DRAU compared to the state-of-the-art on the VQA 1.0 dataset. N corresponds to the number of models used for prediction. WE indicates whether the method uses a pre-trained word embedding. VG indicates whether the method uses external data from the Visual Genome dataset.}
  	\label{tab:vqa1_test}
  	\begin{tabular}{lccccccccccc}
  		\toprule
  		\multicolumn{12}{c}{VQA 1.0 Open Ended Task} \\
  		\midrule
  		\multicolumn{4}{c}{}& \multicolumn{4}{c}{Test-dev} & \multicolumn{4}{c}{Test-standard}  \\
  		
  		Model                                  & N & WE         & VG         & All            & Y/N            & Num.           & Other          & All            & Y/N            & Num.           & Other          \\

  		\cmidrule(r){1-1}   \cmidrule{2-2} \cmidrule{3-4} \cmidrule(l){5-5}  \cmidrule(l){6-8} \cmidrule(l){9-9}  \cmidrule(l){10-12}
  		
  		DMN+ \citep{xiong2016a}                & 1 & -          & -          & 60.3           & 80.5           & 36.8           & 48.3           & 60.4           & -              & -              & -              \\
  		
  		HieCoAtt \citep{lu2016}                & 1 & -          & -          & 61.8           & 79.7           & 38.7           & 51.7           & 62.1           & -              & -              & -              \\
  		RAU \citep{noh2016c}                   & 1 & -          & -          & 63.3           & 81.9           & 39.0           & 53.0           & 63.2           & 81.7           & 38.2           & 52.8           \\
  		DAN \citep{nam2016b}                   & 1 & -          & -          & 64.3           & 83.0           & 39.1           & 53.9           & 64.2           & 82.8           & 38.1           & 54.0           \\
  		MCB \citep{fukui2016}                  & 7 & \checkmark & \checkmark & 66.7           & 83.4           & 39.8           & 58.5           & 66.47          & 83.24          & 39.47          & 58.00          \\ %
  		MLB \citep{kim2016b}                   & 1 & \checkmark & \xmark     & -              & -              & -              & -              & 65.07          & 84.02          & 37.90          & 54.77          \\ %
  		MLB \citep{kim2016b}                   & 7 & \checkmark & \checkmark & 66.77          & 84.57          & 39.21          & 57.81          & 66.89          & 84.61          & 39.07          & 57.79          \\
  		MUTAN \citep{ben-younes2017})          & 5 & \checkmark & \checkmark & 67.42          & \textbf{85.14} & 39.81          & 58.52          & 67.36          & 84.91          & \textbf{39.79} & 58.35          \\ %
  		MFH \citep{yu2017}                     & 1 & \checkmark & \checkmark & \textbf{67.7}  & 84.9           & \textbf{40.2}  & \textbf{59.2}  & \textbf{67.5}  & \textbf{84.91} & 39.3           & \textbf{58.7 } \\ %
  		\cmidrule(r){1-1}   \cmidrule{2-2} \cmidrule{3-4} \cmidrule(l){5-5}  \cmidrule(l){6-8} \cmidrule(l){9-9}  \cmidrule(l){10-12}
  		
  		DRAU\textsubscript{FRCNN + MCB fusion} & 1 & \checkmark & \xmark     & \textbf{66.86} & \textbf{84.92} & \textbf{39.16} & \textbf{57.70} & \textbf{67.16} & \textbf{84.87} & \textbf{40.02} & \textbf{57.91} \\ %
  		\bottomrule
  	\end{tabular}

  \end{table*}
  
  \paragraph{VQA 2.0}
  
  The first place submission \cite{anderson2017} reports using an ensemble of 30 models. In their report, the best single model that also uses FRCNN features achieves $65.67\%$ on the \textit{test-standard} split which is outperformed by our single model (DRAU).
  
  Recently, the VQA 2018 challenge results have been released. It uses the same dataset as the previous VQA 2017 challenge (VQA 2.0). While we have not participated in this challenge, we include the challenge winners results \citep{jiang2018} for the sake of completeness. \citeauthor{jiang2018} builds upon the VQA 2017 challenge winners model by proposing a number of modifications. First, they use weight normalization and ReLU instead of gated hyperbolic tangent activation. For the learning schedule,  the Adam optimizer was swapped for Adamax with a warm up strategy. Moreover, the Faster-RCNN features have been replaced by the state-of-the-art Feature Pyramid Networks (FPN) object detectors. Lastly, they use more additional training data from the common Visual Genome and the new Visual Dialog (VisDial) datasets.
  
  \begin{table*}[t]
  	\center
  	\caption[DRAU compared to the current submissions on the VQA 2.0 dataset]{DRAU compared to the current submissions on the VQA 2.0 dataset. N corresponds to the number of models used for prediction. WE indicates whether the method uses a pre-trained word embedding. VG indicates whether the method uses external data from the Visual Genome dataset.}

  	\begin{tabular}{lccccccccccc}
  		\toprule
  		\multicolumn{12}{c}{VQA 2.0 Open Ended Task} \\
  		\midrule[0.8pt]
  		\multicolumn{4}{c}{}& \multicolumn{4}{c}{Test-dev} & \multicolumn{4}{c}{Test-standard}  \\
  		
  		Model                                  & N  & WE         & VG         & All            & Y/N            & Num.           & Other          & All            & Y/N             & Num.           & Other          \\

  		\cmidrule(r){1-1}   \cmidrule{2-2} \cmidrule{3-4} \cmidrule(l){5-5}  \cmidrule(l){6-8} \cmidrule(l){9-9}  \cmidrule(l){10-12}
  		
  		VQATeam\_MCB \citep{goyal2016}         & 1  & \checkmark & \checkmark & 61.96          & 78.41          & 38.81          & 53.23          & 62.27          & 78.82           & 38.28          & 53.36          \\ %

  		UPMC-LIP6 \citep{ben-younes2017}       & 5  & \checkmark & \checkmark & 65.57          & 81.96          & 41.62          & 57.07          & 65.71          & 82.07           & 41.06          & 57.12          \\ %

  		HDU-USYD-UNCC \citep{yu2017}           & 9  & \checkmark & \checkmark & 68.02          & 84.39          & 45.76          & 59.14          & 68.09          & 84.5            & 45.39          & 59.01          \\ %
  		Adelaide-Teney \citep{teney2017}       & 1  & \checkmark & \checkmark & \textbf{65.32} & \textbf{81.82} & \textbf{44.21} & \textbf{56.05} & \textbf{65.67} & \textbf{82.20}  & \textbf{43.90} & \textbf{56.26} \\ %
  		Adelaide-Teney \citep{anderson2017}    & 30 & \checkmark & \checkmark & -              & -              & -              & -              & 70.34          & 86.60           & 48.64          & 61.15          \\ %
  		FAIR A-STAR \citep{jiang2018}          & 1  & \checkmark & \checkmark & 70.01          & -              & -              & -              & 70.24          & -               & -              & -              \\ %
  		FAIR A-STAR \citep{jiang2018}          & 30 & \checkmark & \checkmark & 72.12          & 87.82          & 51.54          & 63.41          & 72.25          & 87.82           & 51.59          & 63.43          \\ %
  		\cmidrule(r){1-1}   \cmidrule{2-2} \cmidrule{3-4} \cmidrule(l){5-5}  \cmidrule(l){6-8} \cmidrule(l){9-9}  \cmidrule(l){10-12}
  		
  		DRAU\textsubscript{FRCNN + MCB fusion} & 1  & \checkmark & \xmark     & \textbf{66.45} & \textbf{82.85} & \textbf{44.78} & \textbf{57.4}  & \textbf{66.85} & \textbf{ 83.35} & \textbf{44.37} & \textbf{57.63} \\ %
  		\bottomrule
  	\end{tabular}
  	
  	\label{tab:mcdb_rvau}
  \end{table*}
  
  \subsection{Discussion}\label{sec:qualitative}
  
\begin{figure*}
  	\fbox{
  		\resizebox{1.33\columnwidth}{!}{%
  			\begin{tabular}{c c}
  				\begin{tabular}{>{\centering\bfseries}m{1in} >{\centering}m{1in} >{\centering\arraybackslash}m{1in}}
  					\includegraphics[height = 2.5cm, width=\linewidth]{lantern_orig.jpg}  &
  					
  					\frame{\includegraphics[height = 2.5cm, width=\linewidth]{lantern_drau1.png}} & \parbox{2.5cm}{\centering How \colorbox{red!0!}{\strut many} \colorbox{red!32!}{\strut lanterns} \colorbox{red!49!}{\strut hang} \colorbox{red!14!}{\strut off} \colorbox{red!1!}{\strut the} \colorbox{red!1!}{\strut clock} \colorbox{red!3!}{\strut tower}?\\DRAU: 4}\\
  					
  					\parbox{2.5cm}{\centering How many lanterns hang off the clock tower? \\ GT: 4} & \frame{\includegraphics[height = 2.5cm, width=\linewidth]{lantern_mcb0.png}} & MCB: 1
  				\end{tabular}
  				
  				\begin{tabular}{>{\centering\bfseries}m{1in} >{\centering}m{1in} >{\centering\arraybackslash}m{1in}}
  					\includegraphics[height = 2.5cm, width=\linewidth]{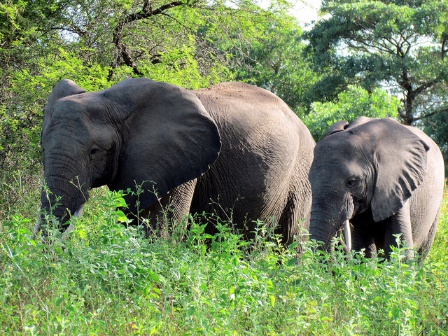}  &
  					
  					\frame{\includegraphics[height = 2.5cm, width=\linewidth]{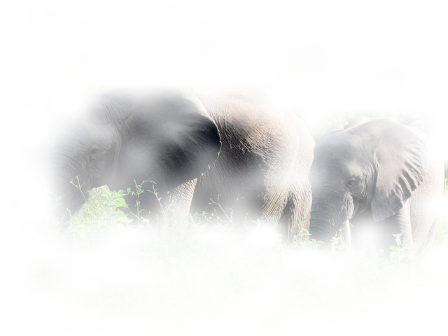}} & \parbox{2.5cm}{\centering How many \colorbox{red!67!}{\strut camels} \colorbox{red!25!}{\strut are} \colorbox{red!4!}{\strut in} \colorbox{red!3!}{\strut the} photo?\\DRAU: 0}\\
  					
  					\parbox{2.5cm}{\centering How many camels are in the photo? \\GT: 0} & \frame{\includegraphics[height = 2.5cm, width=\linewidth]{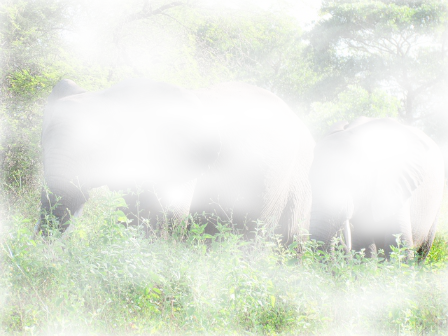}} & MCB: 1
  				\end{tabular}
  			\end{tabular}
  		}
  		\resizebox{0.66\columnwidth}{!}{%
  			
  			\begin{tabular}{>{\centering\bfseries}m{1in} >{\centering}m{1in} >{\centering\arraybackslash}m{1in}}
  				\includegraphics[height = 2.5cm, width=\linewidth]{racket_orig.jpg}  &
  				
  				\frame{\includegraphics[height = 2.5cm, width=\linewidth]{racket_drau1.png}} & \parbox{2.5cm}{\centering What is on the floor leaning on \colorbox{red!8!}{\strut the} \colorbox{red!41!}{\strut bench} \colorbox{red!11!}{\strut in} \colorbox{red!1!}{\strut between} \colorbox{red!8!}{\strut the} \colorbox{red!31!}{\strut people}?\\DRAU: racket}\\
  				
  				\parbox{2.5cm}{\centering What is on the floor leaning on the bench in between the people? \\GT: racket} &\frame{\includegraphics[height = 2.5cm, width=\linewidth]{racket_mcb0.png}} & MCB: backpack
  			\end{tabular}
  		}
  	}
  	\caption{DRAU vs. MCB Qualitative examples. Attention maps for both models shown. DRAU shows subjectively better attention map quality.
  	}
  	\label{fig:count}
  \end{figure*}

  \paragraph{DRAU versus MCB}
  The strength of RAUs is notable in tasks that require sequentially processing the image or relational/multi-step reasoning.  \cref{fig:count} shows some qualitative results between DRAU and MCB. For fair comparison we compare the first attention map of MCB with the second attention map of our model. We do so because the authors of MCB \citep{fukui2016} visualize the first map in their work\footnote[3]{\url{https://github.com/akirafukui/vqa-mcb/blob/master/server/server.py\#L185}}. Furthermore, the first glimpse of our model seems to be the complement of the second attention, i.e.\ the model separates the background and the target object(s) into separate attention maps (illustrated in \cref{fig:vqanet}).
  
  \definecolor{nublue}{RGB}{142, 196, 249}%
  \definecolor{nured}{RGB}{251, 105, 105}%

  \begin{figure}
  	\centering
  	\includegraphics[height = 2.5cm, width =2.5cm]{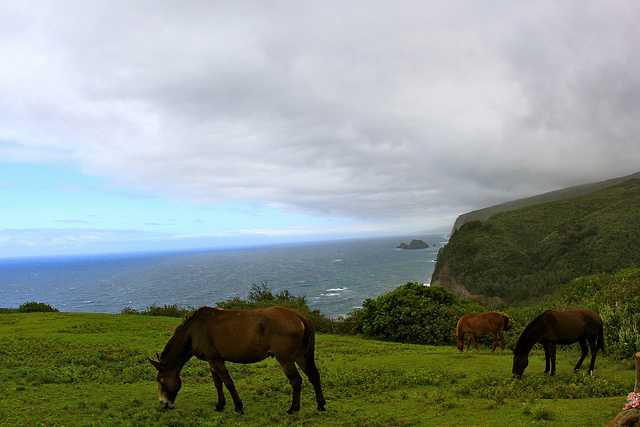}
  	
  	\resizebox{\columnwidth}{!}{%
  		
  		\begin{tabular}{c c}
  			\begin{tabular}{>{\centering\bfseries}m{0in} >{\centering}m{1in} >{\centering\arraybackslash}m{1in}}
  				&
  				
  				\frame{\includegraphics[height = 2.5cm, width=\linewidth]{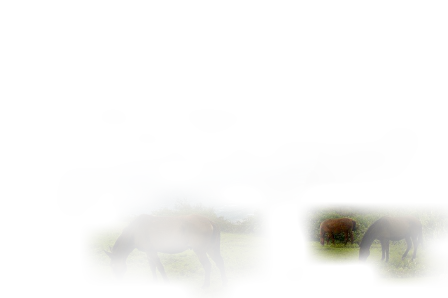}} & \parbox{2.5cm}{\centering How \colorbox{red!31!}{\strut many} \colorbox{red!66!}{\strut horses} \colorbox{red!21!}{\strut are} \colorbox{red!10!}{\strut there} ?\\DRAU: \textbf{3}}\\
  				
  				& \frame{\includegraphics[height = 2.5cm, width=\linewidth]{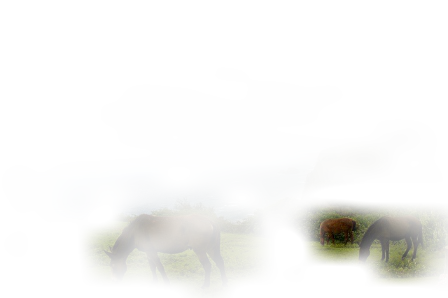}} & \parbox{2.5cm}{\centering How \colorbox{red!31!}{\strut many} \colorbox{red!99!}{\strut dogs} \colorbox{red!21!}{\strut are} \colorbox{red!10!}{\strut there} ? \\DRAU: \textbf{0}}
  			\end{tabular}
  			
  			\begin{tabular}{>{\centering\bfseries}m{1in} >{\centering}m{0in} >{\centering\arraybackslash}m{1in}}
  				\frame{\includegraphics[height = 2.5cm, width=\linewidth]{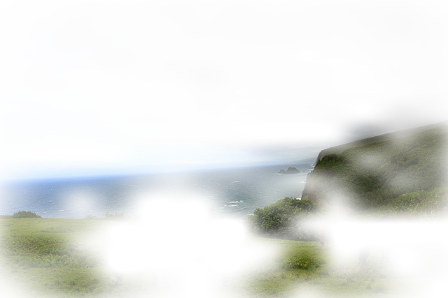}}  &
  				
  				& \parbox{2.5cm}{\centering \colorbox{red!31!}{\strut Where} \colorbox{red!5!}{\strut are} \colorbox{red!5!}{\strut the} \colorbox{red!99!}{\strut horses} ?\\DRAU: \textbf{field}}\\
  				
  				\frame{\includegraphics[height = 2.5cm, width=\linewidth]{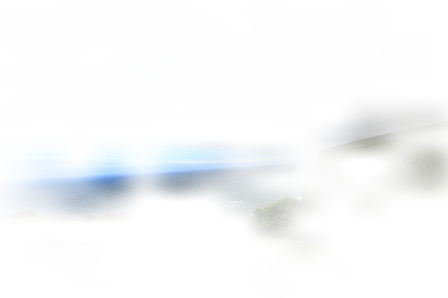}} &   & \parbox{2.5cm}{\centering \colorbox{red!45!}{\strut Can} \colorbox{red!3!}{\strut you} \colorbox{red!37!}{\strut see} \colorbox{red!10!}{\strut the}  \colorbox{red!99!}{\strut horizon} ? \\DRAU: \textbf{yes}}
  			\end{tabular}
  		\end{tabular}
  	}

  	\caption{Four real example results of our proposed model for a single random image. The visual attention, textual attention, and answer are shown. Even on the same image, our model shows rich reasoning capabilities for different question types. The first column shows that the model is able to do two-hop reasoning, initially identifying the  animal in the question and then proceed to correctly count it in the image. The second column results highlights the model's ability to shift its attention to the relevant parts of the image and question. It is worth noting that all the keywords in the questions have the highest attention weights. }
  	\label{fig:vqa_horses}
  \end{figure}

  In \cref{fig:count}, it is clear that the recurrence helps the model attend to multiple targets as apparent in the difference of the attention maps between the two models. The first example shows that \ac{DRAU} attends to the right object (lanterns) both visually and textually. The model also attends to the cars' rear lights as possible ``lanterns'', but the text attention attends to the word ``hang'' which disqualifies the car lights and guides the model to count hanging lanterns. Furthermore, DRAU can predict non-existing object(s). The second example in \cref{fig:count} illustrates that DRAU is not easily fooled by counting whatever animal is present in the image but rather the ``camels'' that is needed to answer the question. This property also translates to questions that require relational reasoning. The third example in \cref{fig:count} demonstrates how well DRAU can attend to the relative location required to answer the question based on the textual and visual attention maps compared to MCB.
  
  \vspace{-.76 mm}
  \paragraph{Attention Quality}\cref{fig:vqa_horses} shows the model's prediction as well as its attention maps for four questions on the same image. It highlights how DRAU can shift the attention intelligently based on different multi-step reasoning questions. To answer the two leftmost questions, a VQA model needs to sequentially process the image and question. First, the model filters out the animals in the picture. Then, the animal in the question is matched to the visual features and finally counted. The two right-most attention maps give a glimpse on how the model filters out the irrelevant parts in the input. Interestingly, inspecting the visual attention for the top question might indicate a bias in the VQA model. Even though the question asks about ``horses'', the visual attention filters out all objects and leaves out the two different backgrounds: sea and field. Since ``horses'' are often found on land, the model predicts ``field'' without any direct attention on the horses in the image.
  
  \section{Conclusion}
  We proposed an architecture for VQA with a recurrent attention mechanism, termed the \acf{RAU}. The recurrent layers help guide the textual and visual attention since the network can reason relations between several parts of the image and question. We provided quantitative and qualitative results indicating the usefulness of a recurrent attention mechanism. Using a simple VQA baseline, we have shown the performance advantage of recurrent attention compared to the traditional convolutional attention used in most VQA models. Furthermore, we performed an ablation study with all possible combinations of recurrent and convolutional attention. The results of the study indicate that recurrent attention can be very beneficial in dual attention VQA models. 
  	Then, we demonstrated that substituting the visual attention mechanism in other networks, MCB \citep{fukui2016}, MUTAN \citep{ben-younes2017}, and MFH \citep{yu2017}, consistently improves their performance.  In VQA 1.0, we come a very close second to the state-of-the-art model. While using the same image features, our DRAU network outperforms the VQA 2017 challenge winner \cite{anderson2017} in a single-model scenario.
  
  In future work we will investigate implicit recurrent attention mechanism using recently proposed explanation methods \citep{ArrWASSA17, MonDSP18} and explore different recurrent models for attention \citep{kalchbrenner2015,shi2015}.
  
  \section*{Acknowledgments}
  This work was supported by the Fraunhofer Society through the MPI-FhG collaboration project "Theory and Practice for Reduced Learning Machines". This research was also supported by the German Ministry for Education and Research as Berlin Big Data Center under Grant 01IS14013A and the Berlin Center for Machine Learning under Grant 01IS180371.


  	\bibliographystyle{elsarticle-harv}
  	\bibliography{vqa_ref}

\begin{thebibliography}{40}
\expandafter\ifx\csname natexlab\endcsname\relax\def\natexlab#1{#1}\fi
\expandafter\ifx\csname url\endcsname\relax
  \def\url#1{\texttt{#1}}\fi
\expandafter\ifx\csname urlprefix\endcsname\relax\def\urlprefix{URL }\fi

\bibitem[{Anderson et~al.(2017)Anderson, He, Buehler, Teney, Johnson, Gould,
  and Zhang}]{anderson2017}
Anderson, P., He, X., Buehler, C., Teney, D., Johnson, M., Gould, S., Zhang,
  L., 2017. Bottom-{{Up}} and {{Top}}-{{Down Attention}} for {{Image
  Captioning}} and {{VQA}}. arXiv:1707.07998.

\bibitem[{Antol et~al.(2015)Antol, Agrawal, Lu, Mitchell, Zitnick, Batra, and
  Parikh}]{antol2015}
Antol, S., Agrawal, A., Lu, J., Mitchell, M., Zitnick, C.~L., Batra, D.,
  Parikh, D., 2015. {{VQA}}: {{Visual Question Answering}}. In: {CVPR}. pp.
  2425--2433.

\bibitem[{Arras et~al.(2017)Arras, Montavon, M{\"u}ller, and
  Samek}]{ArrWASSA17}
Arras, L., Montavon, G., M{\"u}ller, K.-R., Samek, W., 2017. Explaining
  {{Recurrent Neural Network Predictions}} in {{Sentiment Analysis}}. In:
  {{EMNLP}}'17 {{Workshop}} on {{Computational Approaches}} to
  {{Subjectivity}}, {{Sentiment}} \& {{Social Media Analysis}} ({{WASSA}}). pp.
  159--168.

\bibitem[{Bahdanau et~al.(2015)Bahdanau, Cho, and Bengio}]{bahdanau2014}
Bahdanau, D., Cho, K., Bengio, Y., 2015. Neural {{Machine Translation}} by
  {{Jointly Learning}} to {{Align}} and {{Translate}}. In: {ICLR}.

\bibitem[{Ben-younes et~al.(2017)Ben-younes, Cadene, Cord, and
  Thome}]{ben-younes2017}
Ben-younes, H., Cadene, R., Cord, M., Thome, N., 2017. {{MUTAN}}: {{Multimodal
  Tucker Fusion}} for {{Visual Question Answering}}. arXiv:1705.06676.

\bibitem[{Bosse et~al.(2018)Bosse, Maniry, M{\"u}ller, Wiegand, and
  Samek}]{BosTIP18}
Bosse, S., Maniry, D., M{\"u}ller, K.-R., Wiegand, T., Samek, W., 2018. Deep
  {{Neural Networks}} for {{No}}-{{Reference}} and {{Full}}-{{Reference Image
  Quality Assessment}}. IEEE Trans. Image Process. 27~(1), 206--219.

\bibitem[{Charikar et~al.(2004)Charikar, Chen, and
  Farach-Colton}]{charikar2004}
Charikar, M., Chen, K., Farach-Colton, M., 2004. Finding frequent items in data
  streams. Theor. Comput. Sci. 312~(1), 3--15.

\bibitem[{Cho et~al.(2014)Cho, {van Merrienboer}, Gulcehre, Bahdanau, Bougares,
  Schwenk, and Bengio}]{cho2014a}
Cho, K., {van Merrienboer}, B., Gulcehre, C., Bahdanau, D., Bougares, F.,
  Schwenk, H., Bengio, Y., Jun. 2014. Learning {{Phrase Representations}} using
  {{RNN Encoder}}-{{Decoder}} for {{Statistical Machine Translation}}.
  arXiv:1406.1078 [cs, stat].

\bibitem[{Fukui et~al.(2016)Fukui, Park, Yang, Rohrbach, Darrell, and
  Rohrbach}]{fukui2016}
Fukui, A., Park, D.~H., Yang, D., Rohrbach, A., Darrell, T., Rohrbach, M.,
  2016. Multimodal {{Compact Bilinear Pooling}} for {{Visual Question
  Answering}} and {{Visual Grounding}}. In: {EMNLP}. pp. 457--468.

\bibitem[{Gao et~al.(2016)Gao, Beijbom, Zhang, and Darrell}]{gao2016}
Gao, Y., Beijbom, O., Zhang, N., Darrell, T., 2016. Compact bilinear pooling.
  In: {CVPR}. pp. 317--326.

\bibitem[{Glorot and Bengio(2010)}]{glorot2010}
Glorot, X., Bengio, Y., 2010. Understanding the {{Difficulty}} of {{Training
  Deep Feedforward Neural Networks}}. In: {{AISTATS}}. pp. 249--256.

\bibitem[{Goyal et~al.(2017)Goyal, Khot, Summers-Stay, Batra, and
  Parikh}]{goyal2016}
Goyal, Y., Khot, T., Summers-Stay, D., Batra, D., Parikh, D., 2017. Making the
  {{V}} in {{VQA Matter}}: {{Elevating}} the {{Role}} of {{Image
  Understanding}} in {{Visual Question Answering}}. In: {CVPR}. pp. 6904--6913.

\bibitem[{He et~al.(2015)He, Zhang, Ren, and Sun}]{he2015}
He, K., Zhang, X., Ren, S., Sun, J., 2015. Delving {{Deep}} into
  {{Rectifiers}}: {{Surpassing Human}}-{{Level Performance}} on {{ImageNet
  Classification}}. In: {ICCV}. pp. 1026--1034.

\bibitem[{Homayounfar et~al.(2018)Homayounfar, Ma, Lakshmikanth, and
  Urtasun}]{homayounfar2018}
Homayounfar, N., Ma, W., Lakshmikanth, S.~K., Urtasun, R., Jun. 2018.
  Hierarchical {{Recurrent Attention Networks}} for {{Structured Online Maps}}.
  In: 2018 {{IEEE}}/{{CVF Conference}} on {{Computer Vision}} and {{Pattern
  Recognition}}. pp. 3417--3426.

\bibitem[{Jiang et~al.(2018)Jiang, Natarajan, Chen, Rohrbach, Batra, and
  Parikh}]{jiang2018}
Jiang, Y., Natarajan, V., Chen, X., Rohrbach, M., Batra, D., Parikh, D., Jul.
  2018. Pythia v0.1: {{The Winning Entry}} to the {{VQA Challenge}} 2018.
  ArXiv180709956 Cs.

\bibitem[{Kalchbrenner et~al.(2015)Kalchbrenner, Danihelka, and
  Graves}]{kalchbrenner2015}
Kalchbrenner, N., Danihelka, I., Graves, A., 2015. Grid long short-term memory.
  arXiv:1507.01526.

\bibitem[{Kim et~al.(2017)Kim, On, Kim, Ha, and Zhang}]{kim2016b}
Kim, J.-H., On, K.-W., Kim, J., Ha, J.-W., Zhang, B.-T., 2017. Hadamard
  {{Product}} for {{Low}}-{{Rank Bilinear Pooling}}. In: {ICLR}.

\bibitem[{Kingma and Ba(2014)}]{kingma2014}
Kingma, D.~P., Ba, J., 2014. Adam: {{A Method}} for {{Stochastic
  Optimization}}. arXiv:1412.6980.

\bibitem[{Krishna et~al.(2017)Krishna, Zhu, Groth, Johnson, Hata, Kravitz,
  Chen, Kalantidis, Li, Shamma, and {others}}]{krishna2017}
Krishna, R., Zhu, Y., Groth, O., Johnson, J., Hata, K., Kravitz, J., Chen, S.,
  Kalantidis, Y., Li, L.-J., Shamma, D.~A., {others}, 2017. Visual genome:
  {{Connecting}} language and vision using crowdsourced dense image
  annotations. Int. J. Comput. Vis. 123~(1), 32--73.

\bibitem[{Kumar and Varaiya(2015)}]{kumar2015}
Kumar, P.~R., Varaiya, P., 2015. Stochastic Systems: {{Estimation}},
  Identification, and Adaptive Control. {SIAM}.

\bibitem[{Lanctot et~al.(2017)Lanctot, Zambaldi, Gruslys, Lazaridou, Tuyls,
  Perolat, Silver, and Graepel}]{lanctot2017}
Lanctot, M., Zambaldi, V., Gruslys, A., Lazaridou, A., Tuyls, K., Perolat, J.,
  Silver, D., Graepel, T., Nov. 2017. A {{Unified Game}}-{{Theoretic Approach}}
  to {{Multiagent Reinforcement Learning}}. arXiv:1711.00832 [cs].

\bibitem[{Lin et~al.(2014)Lin, Maire, Belongie, Hays, Perona, Ramanan,
  Doll{\'a}r, and Zitnick}]{lin2014}
Lin, T.-Y., Maire, M., Belongie, S., Hays, J., Perona, P., Ramanan, D.,
  Doll{\'a}r, P., Zitnick, C.~L., 2014. Microsoft {{Coco}}: {{Common Objects}}
  in {{Context}}. In: {ECCV}. pp. 740--755.

\bibitem[{Lu et~al.(2016)Lu, Yang, Batra, and Parikh}]{lu2016}
Lu, J., Yang, J., Batra, D., Parikh, D., 2016. Hierarchical
  {{Question}}-{{Image Co}}-{{Attention}} for {{Visual Question Answering}}.
  In: {NIPS}. pp. 289--297.

\bibitem[{Mnih et~al.(2014)Mnih, Heess, Graves, and Kavukcuoglu}]{mnih2014}
Mnih, V., Heess, N., Graves, A., Kavukcuoglu, K., 2014. Recurrent {{Models}} of
  {{Visual Attention}}. In: Proceedings of the 27th {{International
  Conference}} on {{Neural Information Processing Systems}} - {{Volume}} 2.
  NIPS'14. {MIT Press}, Cambridge, MA, USA, pp. 2204--2212.

\bibitem[{Montavon et~al.(2018)Montavon, Samek, and M{\"u}ller}]{MonDSP18}
Montavon, G., Samek, W., M{\"u}ller, K.-R., 2018. Methods for {{Interpreting}}
  and {{Understanding Deep Neural Networks}}. Digit. Signal Process. 73, 1--15.

\bibitem[{Nallapati et~al.(2016)Nallapati, Zhou, Gulcehre, Xiang, and
  {others}}]{nallapati2016abstractive}
Nallapati, R., Zhou, B., Gulcehre, C., Xiang, B., {others}, 2016. Abstractive
  text summarization using sequence-to-sequence rnns and beyond.
  arXiv:1602.06023.

\bibitem[{Nam et~al.(2017)Nam, Ha, and Kim}]{nam2016b}
Nam, H., Ha, J.-W., Kim, J., 2017. Dual {{Attention Networks}} for {{Multimodal
  Reasoning}} and {{Matching}}. In: {CVPR}. pp. 299--307.

\bibitem[{Noh and Han(2016)}]{noh2016c}
Noh, H., Han, B., 2016. Training {{Recurrent Answering Units}} with {{Joint
  Loss Minimization}} for {{VQA}}. arXiv:1606.03647.

\bibitem[{Paulus et~al.(2017)Paulus, Xiong, and Socher}]{paulus2017}
Paulus, R., Xiong, C., Socher, R., May 2017. A {{Deep Reinforced Model}} for
  {{Abstractive Summarization}}. arXiv:1705.04304 [cs].

\bibitem[{Pennington et~al.(2014)Pennington, Socher, and
  Manning}]{pennington2014glove}
Pennington, J., Socher, R., Manning, C.~D., 2014. {{GloVe}}: {{Global Vectors}}
  for {{Word Representation}}. In: {EMNLP}. pp. 1532--1543.

\bibitem[{Ren et~al.(2015)Ren, He, Girshick, and Sun}]{ren2015b}
Ren, S., He, K., Girshick, R., Sun, J., Jun. 2015. Faster {{R}}-{{CNN}}:
  {{Towards Real}}-{{Time Object Detection}} with {{Region Proposal Networks}}.
  arXiv:1506.01497.

\bibitem[{Rensink(2000)}]{rensink2000}
Rensink, R.~A., Jan. 2000. The {{Dynamic Representation}} of {{Scenes}}. Vis.
  Cogn. 7~(1-3), 17--42.

\bibitem[{Schwartz et~al.(2017)Schwartz, Schwing, and Hazan}]{schwartz2017}
Schwartz, I., Schwing, A.~G., Hazan, T., Nov. 2017. High-{{Order Attention
  Models}} for {{Visual Question Answering}}. arXiv:1711.04323 [cs].

\bibitem[{Shi et~al.(2015)Shi, Chen, Wang, Yeung, Wong, and Woo}]{shi2015}
Shi, X., Chen, Z., Wang, H., Yeung, D.-Y., Wong, W.-k., Woo, W.-c., Jun. 2015.
  Convolutional {{LSTM Network}}: {{A Machine Learning Approach}} for
  {{Precipitation Nowcasting}}. arXiv:1506.04214 [cs].

\bibitem[{Teney et~al.(2017)Teney, Anderson, He, and {van den
  Hengel}}]{teney2017}
Teney, D., Anderson, P., He, X., {van den Hengel}, A., 2017. Tips and
  {{Tricks}} for {{Visual Question Answering}}: {{Learnings}} from the 2017
  {{Challenge}}. arXiv:1708.02711.

\bibitem[{Tucker(1966)}]{tucker1966}
Tucker, L.~R., 1966. Some {{Mathematical Notes}} on {{Three}}-{{Mode Factor
  Analysis}}. Psychometrika 31~(3), 279--311.

\bibitem[{Xiong et~al.(2016)Xiong, Merity, and Socher}]{xiong2016a}
Xiong, C., Merity, S., Socher, R., 2016. Dynamic {{Memory Networks}} for
  {{Visual}} and {{Textual Question Answering}}. In: {ICML}. pp. 2397--2406.

\bibitem[{Xu and Saenko(2016)}]{xu2015}
Xu, H., Saenko, K., 2016. Ask, {{Attend}} and {{Answer}}: {{Exploring
  Question}}-{{Guided Spatial Attention}} for {{Visual Question Answering}}.
  In: {ECCV}. pp. 451--466.

\bibitem[{Yu et~al.(2017{\natexlab{a}})Yu, Yu, Fan, and Tao}]{yu2017a}
Yu, Z., Yu, J., Fan, J., Tao, D., Aug. 2017{\natexlab{a}}. Multi-{{Modal
  Factorized Bilinear Pooling}} with {{Co}}-{{Attention Learning}} for {{Visual
  Question Answering}}. ArXiv170801471 Cs.

\bibitem[{Yu et~al.(2017{\natexlab{b}})Yu, Yu, Xiang, Fan, and Tao}]{yu2017}
Yu, Z., Yu, J., Xiang, C., Fan, J., Tao, D., 2017{\natexlab{b}}. Beyond
  {{Bilinear}}: {{Generalized Multi}}-{{Modal Factorized High}}-{{Order
  Pooling}} for {{Visual Question Answering}}. arXiv:1708.03619.

\end{thebibliography}
  	
  \end{document}